\documentclass[11pt,a4paper]{article}
\usepackage[]{acl}
\usepackage{latexsym}
\usepackage{arydshln}
\usepackage{soul}

\usepackage{microtype}

\usepackage{array}
\usepackage{pifont}
\usepackage{tabularx}
\usepackage{adjustbox}
\usepackage{multirow}
\usepackage{enumitem}
\usepackage{xspace}
\usepackage{tcolorbox}
\usepackage{booktabs,amsfonts,dcolumn}
\usepackage{url}
\usepackage{amsmath,amsthm,amsfonts,amssymb,bm,stmaryrd}
\usepackage[noorphans,vskip=0.75ex,leftmargin=2ex]{quoting}

\usepackage{amsmath}%
\usepackage{kantlipsum}
\usepackage{tabularx}
\usepackage[utf8]{inputenc} %
\usepackage[T1]{fontenc}    %
\usepackage{url}            %
\usepackage{booktabs,multirow}       %
\usepackage{amsfonts}       %
\usepackage{nicefrac}       %
\usepackage{microtype}      %

\usepackage{graphicx}
\usepackage{times}
\usepackage{epsfig}
\usepackage{graphicx}
\usepackage{amsmath}
\usepackage{amssymb}
\usepackage{xcolor,vwcol}
\usepackage{booktabs}

\usepackage[OT6,T1]{fontenc}
\usepackage[utf8]{inputenc}
\usepackage[english]{babel}

\usepackage{mathtools}
\DeclarePairedDelimiterX{\norm}[1]{\lVert}{\rVert}{#1}
\usepackage{multicol}
\usepackage[linesnumbered,ruled,vlined]{algorithm2e}
\usepackage{setspace}

\let\oldnl\nl%
\newcommand{\nonl}{\renewcommand{\nl}{\let\nl\oldnl}}%

\DeclareMathAlphabet{\pazocal}{OMS}{zplm}{m}{n} 
\usepackage{microtype}
\usepackage{graphicx}
\usepackage{colortbl}

\usepackage{booktabs} %
\usepackage{subcaption}
\usepackage{times}
\usepackage{latexsym}
\usepackage{makecell} 
\usepackage{amsfonts} 
\usepackage{amsmath} 
\usepackage{amssymb}

\newcommand{\armenian}{\fontencoding{OT6}\fontfamily{cmr}\selectfont}
\DeclareTextFontCommand{\textarmenian}{\armenian}

\definecolor{brightmaroon}{rgb}{0.76, 0.23, 0.28}

\newcommand{\sect}{\S}
\newcommand{\red}[1]{\textcolor{black}{#1}}

\definecolor{cadmiumgreen}{rgb}{0.0, 0.42, 0.24}
\definecolor{burgundy}{rgb}{0.5, 0.0, 0.13}
\definecolor{darkviolet}{rgb}{0.58, 0.0, 0.83}
\definecolor{princetonorange}{rgb}{1.0, 0.56, 0.0}
\definecolor{darkmagenta}{rgb}{0.55, 0.0, 0.55}
\definecolor{cornellred}{rgb}{0.7, 0.11, 0.11}

\renewcommand{\paragraph}[1]{\vspace{0.2cm}\noindent\textbf{#1}}

\DeclareSymbolFont{extraup}{U}{zavm}{m}{n}
\DeclareMathSymbol{\varheart}{\mathalpha}{extraup}{86}
\DeclareMathSymbol{\vardiamond}{\mathalpha}{extraup}{87}

\title{Long Context Question Answering via Supervised Contrastive Learning}

\author{Avi Caciularu$^{\clubsuit}\thanks{\;\; Work partly done as an intern at AI2.}$\hspace{1em}
\textbf{Ido Dagan$^\clubsuit$ \hspace{1em}Jacob Goldberger$^{\spadesuit}$ }\hspace{1em} Arman Cohan$^{\diamondsuit,\heartsuit}$\vspace{6pt}\\  
    $^\clubsuit$Computer Science Department, Bar-Ilan University, Ramat-Gan, Israel\\
    $^\spadesuit$Faculty of Engineering, Bar-Ilan University, Ramat-Gan, Israel\\
    $^\diamondsuit$Allen Institute for AI, Seattle, WA\\
    $^\heartsuit$Paul G. Allen School of Computer Science, University of Washington, Seattle, WA\\
    {\small\tt avi.c33@gmail.com, armanc@allenai.org} \\
    {\small\tt dagan@cs.biu.ac.il, jacob.goldberger@biu.ac.il }
}
\date{}

\begin{document}

\maketitle
\begin{abstract}
Long-context question answering (QA) tasks require reasoning over a long document or multiple documents. Addressing these tasks often benefits from identifying a set of evidence spans (e.g., sentences), which provide supporting evidence for answering the question.
In this work, we propose a novel method for equipping long-context QA models with an additional sequence-level objective for better identification of the supporting evidence.
We achieve this via an additional contrastive supervision signal in finetuning, where the model is encouraged to explicitly discriminate supporting evidence sentences from negative ones by maximizing question-evidence similarity. 
The proposed additional loss exhibits consistent improvements on three different strong long-context transformer models, across two challenging question answering benchmarks -- HotpotQA and QAsper.\footnote{Our code is available at \url{https://github.com/aviclu/long-context-qa-contrast}.}
\end{abstract}

\section{Introduction}
Answering questions that require reasoning over a long sequence, such over long documents or multiple documents, is a challenging task~\cite{quality2021}. 
Research in this domain mostly includes tasks that involve multiple text segments, over benchmarks like HotpotQA~\cite{yang-etal-2018-hotpotqa} and QAsper \cite{dasigi-etal-2021-dataset}. 
HotpotQA is a multi-hop QA benchmark over multiple paragraphs from Wikipedia, while QAsper involves reading comprehension from long academic papers, where relevant information on a question could be spread across the paper. 

Given the task complexity \cite{choi-etal-2017-coarse}, benchmarks often provide an additional set of evidence spans, such as sentences or paragraphs, for a given question answer pair. 
This breaks down the long-context QA task, adding a preliminary evidence span detection, which is crucial for successfully finding the correct answer, and also potentially helps in model interpretability. In this work, we propose a method for improving long-context QA via leveraging such evidence spans, by maximizing their similarity with the question.

\begin{figure*}
\centering
  \centering
\includegraphics[width=0.99\linewidth]{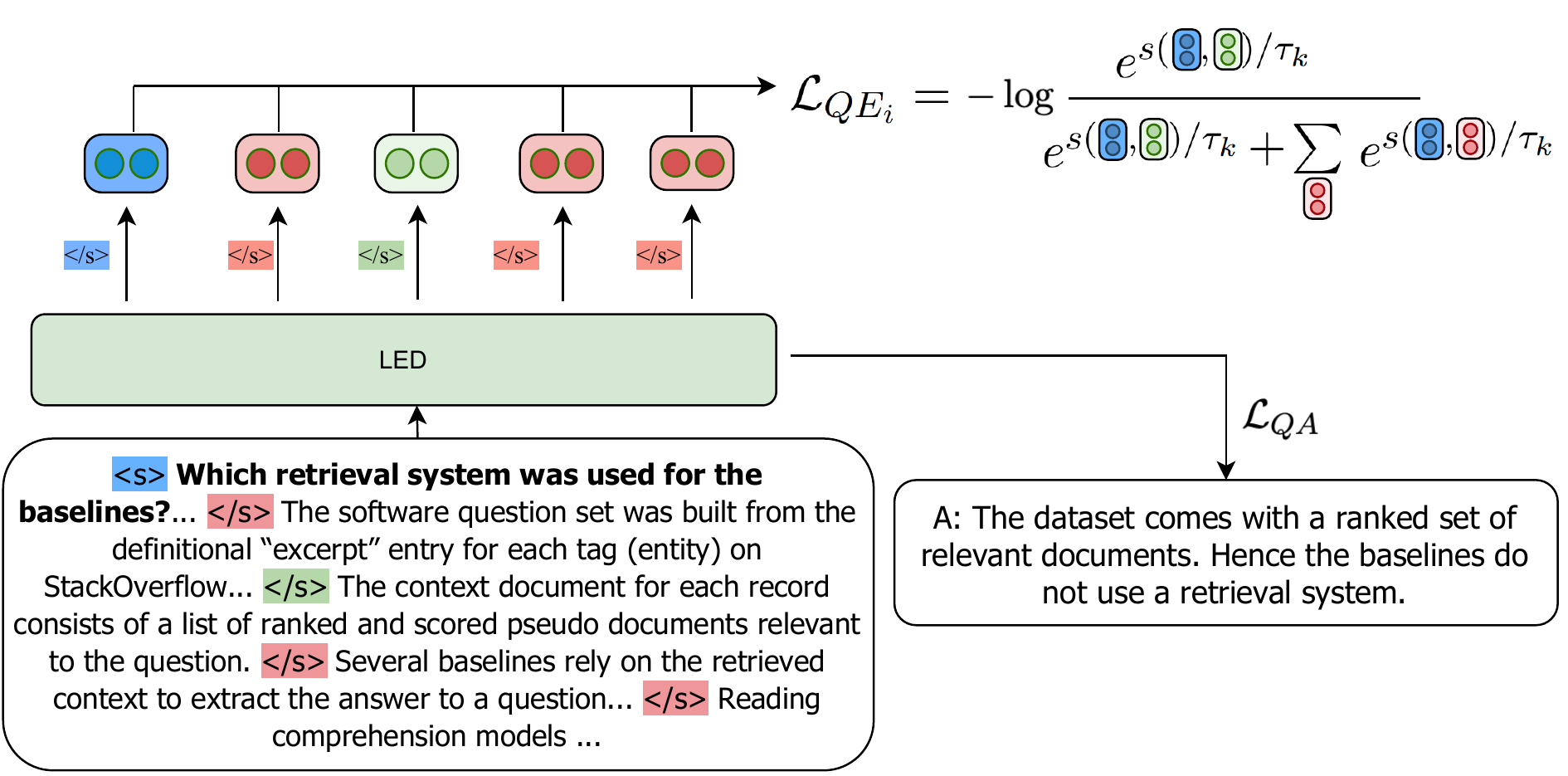}
\caption{Demonstration of our method over an example instance from QAsper~\cite{dasigi-etal-2021-dataset}. A long sequence is fed to a model, producing representations for the marker special tokens. Then, these vectors are used to compute our contrastive objective. The token colored in blue represents the question, and the tokens colored in green (red) represent positive (negative) evidence sentences. The goal is to maximize the similarity between the blue vector and the green vectors.}
\label{fig:model}
\end{figure*}

Since identifying the evidences provides relevant information for answering the question, prior work showed that jointly training models to perform evidence span extraction in addition to answer generation is important for achieving high performance \cite{yang-etal-2018-hotpotqa,dasigi-etal-2021-dataset}. 
To jointly perform evidence extraction and question answering, models utilize sentence representations (marker tokens) in the input; the final layer representation corresponding to these markers is then passed through a classification layer and is optimized using the cross-entropy loss in conjunction with the answer extraction/generation loss. We conjecture and demonstrate (see Table~\ref{tab:map}) that this objective does not sufficiently capture relationships between the question and the candidate evidence spans. Thus, we propose a complementary objective for enforcing question-evidence \textit{similarity} in the model representation (see~Fig.~\ref{fig:model}). Further, we show that optimizing the question-evidence similarity under a certain subspace by using linear projection of the raw representations may softly impose information encoding about their relatedness. Since questions may be partitioned into several types (e.g., yes/no, generative, non answerable, etc.) in the common QA settings, we also investigate learning separate projections per question type.

Driven by the intuition that questions should be similar to their supporting evidences, under some specific geometric sub-space, we propose a novel supervised contrastive learning objective for the finetuning stage, aiming to maximize similarity of question-evidence representations. Contrastive learning has been recently applied to a variety of deep learning models in computer vision \cite{chen2020simple, chen2021exploring} and NLP~\cite{gao-etal-2021-simcse,gunel2021supervised}. Unlike prior NLP related methods, our proposed loss term is model-agnostic and operates in the \textit{sequence-level} in a supervised manner. This objective targets question and evidence representations within input sequences, and unlike prior work, it is not based on individually encoding sentences or paragraphs.  We show that our additional contrastive supervision provides consistent improvements on three different models and two datasets, demonstrating its effectiveness and versatility.

\section{Setup}
\label{sec:background}
In this section, we define and elaborate the common setup and notations for long-context QA.

Assume that we are given a question $q_i$ and a context $\mathcal{S}_i{=}\langle {s_1},...,{s_M}\rangle$ consisting of $M$ sentences ($s_j$ can be also a document/paragraph/passage, depending on the dataset). From $\mathcal{S}_i$, the task is to identify the correct answer $a_i$ and a set of $N$ evidence spans $\mathcal{S}^+_i{=}\{s_{i_1},..., s_{i_N}\}$ where $i_j$ are indices of the sentences that are the supporting evidence for answering the question $q_i$. %

As common in the input setup of long-context transformer models \cite{longformer,zaheer2020big,caciularu-etal-2021-cdlm-cross}, which are the current state-of-the-art models for solving long-context QA tasks, 
the question and context sentences are concatenated in a single long sequence with special sentence tokens specifying sentence boundaries. 
Then the input is passed to the long-context transformer, which is trained to jointly identify the evidence sentences and extract or generate the answer.

Concretely, for each example, we prepare the following concatenated input sequence:
\begin{equation*}
[\texttt{$\langle$s$\rangle$},q_i,\texttt{$\langle$/s$\rangle$},s_{1},\texttt{$\langle$/s$\rangle$},s_{2},...,\texttt{$\langle$/s$\rangle$},s_{M}]
\label{eq:input}
\end{equation*}
where ``,'' is the string concatenation operation, $q_i$ and $s_j$ are sequences of tokens corresponding to the question and the $j$th sentence in the input context, and \texttt{$\langle$s$\rangle$} and \texttt{$\langle$/s$\rangle$} are special tokens representing the question and a context sentence, respectively (See Fig.~\ref{fig:model} for an example). %

Then, a QA loss, which we denote by $\mathcal{L}_{QA}$ is applied over the contextualized representation of each sentence token, and is optimized using supervision.  $\mathcal{L}_{QA}$ depends on the dataset and can take the form of a multi-task objective, representing multiple tasks in the context of QA~\cite{dasigi-etal-2021-dataset} (particularly evidence extraction and answer generation).

\section{Question-Evidence Contrastive Loss}
\label{sec:method}
In this section, we elaborate on our novel proposed contrastive loss term, combining question-evidence similarity maximization (\sect\ref{subsec:qe_sim}), and question-type aware projections (\sect\ref{subsec:qt_aware}).

\subsection{Question-Evidence Similarity}
\label{subsec:qe_sim}
To encourage the long-context transformer model to capture relationships between the question and evidence sentences, we introduce an additional sequence-level loss that compares and contrasts the question with context sentences. 

The additional proposed loss $\mathcal{L}_{QE}$ is based on the InfoNCE loss~\cite{oord2018representation}, and is applied over instances consisting of triplet of vectors representing the question, an evidence sentence and distractor sentences. The loss encourages the question and evidence representations to become closer to each other, while pushing the question and distracting sentences away. 

Formally, the contrastive loss is defined as the sum of negative log-likelihood losses over all input examples, where each loss term discriminates the positive units from negative ones:
\begin{align}\label{eqn:loss}
    \mathcal{L}_{QE_i} = -\log \sum\limits_{s^+ \in \mathcal{S}^+_i} \left(\frac{e^{sim(s^+, q_i)/\tau_k}}{\sum\limits_{s \in \mathcal{S}_i} e^{sim(s, q_i)/\tau_k}}\right),
\end{align}
where $s$ ($q_i$) is the sentence (question) marker vector representation (see \texttt{$\langle$s$\rangle$} and \texttt{$\langle$/s$\rangle$} in Fig.~\ref{fig:model})%
, $\tau_k$ is the configurable temperature hyperparameter, and $sim(\cdot)$ is a similarity metric. $\mathcal{L}_{QE_i}$ serves a single example, and the final aggregated loss ${\mathcal{L}_{QE}}$ is obtained by averaging over all the examples.

We incorporate $\mathcal{L}_{QE}$ into the main QA span extraction/generation loss $\mathcal{L}_{QA}$ using the augmented loss:
\begin{equation*}
{\mathcal{L}=(1-\lambda)\cdot\mathcal{L}_{QA}+\lambda\cdot\mathcal{L}_{QE},}
\end{equation*}
where $\lambda$ is a weighting hyperparameter.

The underlying $sim(\cdot)$ can take the form of a non-parametric similarity function, e.g., the dot product ($sim\left(s, q\right){=}s^\top q$) or the cosine similarity ($sim\left(s, q\right){=}\frac{s^\top q}{\norm{s}\norm{q}}$). However, we show empirically that using such similarity over the raw representations harms the performance results of the model, since seemingly, it is hard to find a shared representation that should optimize the two loss functions. Hence, we adopted linear projections, per question type, to cast the similarity learning objective into proxy linear spaces.

\subsection{Incorporating Question-Type Projections}
\label{subsec:qt_aware}
\red{Long-context QA benchmarks often provide a question-type label per instance as an additional challenge, such as ${\text{\{yes, no, span\}}}$ for HotpotQA. We hypothesize that maximizing question-evidence similarity under a question-type-specific sub-space can enable more flexibility and inductive bias to the model, for producing better representations and further improving the performance.}
Following \citet{iter-etal-2020-pretraining}, we define the following similarity function:
\begin{equation}
sim_k\left(s,q\right)=s^\top W_k q,
\label{eq:first_s}
\end{equation}
where $k$ is the expected question type and $W_k$ is the corresponding learnable projection matrix. Such linear projections ensure that a specific subspace per question type exists. We additionally incorporate different temperature hyperparameters $\tau_k$ per question type in Eq.~\ref{eqn:loss} (see the ablation in Table \ref{tab:ablations} for their effect). %
The dimensions of the proposed $W_k$ tend to be large, in accordance with the dimensions of the transformer's hidden-layers.\footnote{Empirically, we end up with $W_k\in\mathbb{R}^{768\times768}$ for base-sized models and $W_k\in\mathbb{R}^{1024 \times1024}$ for large-sized models.} Hence, following~\citet{barkan-etal-2020-within}, we apply new non-square linear projections instead of using $W_k$:
\begin{equation}
sim_k\left(s,q\right)=\frac{s_k^\top q_k}{\norm{s_k}\norm{q_k}},
\label{eq:cos_k}
\end{equation}
where we set $$s_k{:=}W^S_{k}s,\quad
q_k{:=}W^Q_{k}q,$$ and
$W^S_{k}$ (or $W^Q_{k}$) is the matrix that projects $s$ (or $q$) into a lower dimension, in the $k^\text{th}$ question-type sub-space. 
In order to improve the separation between the different sub-spaces induced by different question types, we generated additional negative instances per sentence, as follows. 

We projected every question-sentence pair using all the mappings according to the available question types, and computed their cosine similarity (according to Eq.~\ref{eq:cos_k}). Then, all the obtained scores were considered as negative, except the ones that belong to question-evidence pairs projected using the correct question-type mapping. %

\begin{table}[t]
\centering
\footnotesize
\setlength{\tabcolsep}{3pt} 
        \begin{tabular}[b]{@{}clrrr@{}}
            \toprule
            &Method            & Yes & No & Span \\ \midrule
            &Trained only with $\mathcal{L}_{QA}$ & 8.4   & 4.9    & 46.4     \\ 
            & $+\mathcal{L}_{QE}$ using Eq.~\ref{eq:cos_k} with a single projection & 72.4 & 65.2 & 77.5\\
            &$+\mathcal{L}_{QE}$ using Eq.~\ref{eq:cos_k} & 86.3 & 84.2 & 87.6\\
            \bottomrule
        \end{tabular}
\caption{Mean Average Precision (mAP) question-evidence cosine similarity results per question type for the HotpotQA-distractor dev set.} 
\label{tab:map}
\end{table}
\begin{table*}[t]
\footnotesize
\centering
\renewcommand{\arraystretch}{1.}
\begin{tabular}{@{}lcccccccc|cc|cr@{}}
\toprule
\multirow{2}{*}{Input} & \multicolumn{2}{c}{Extractive} & \multicolumn{2}{c}{Abstractive} & \multicolumn{2}{c}{Yes/No} & \multicolumn{2}{c}{Unanswerable} &
\multicolumn{2}{c}{Evidence} & \multicolumn{2}{c}{Overall} \\
& Dev. & Test & Dev. & Test & Dev. & Test & Dev. & Test & Dev. & Test & Dev. & Test \\
\midrule
LED     \citeyearpar{longformer}                    & 26.1 & 31.0 & 16.6 & 15.8 & 67.5 & 70.3 & 28.6 & 26.2 & 23.9  & 29.9& 29.1&  32.8\\
+ $\mathcal{L}_{QE}$ ($\Delta$)  & +0.2 & +0.2 & +0.8 & +1.0 & +1.6 & +0.2 & +1.5 & +1.9 & +1.0 & +0.7 & +0.9 & +0.7 \\
\bottomrule
\end{tabular}
\caption{Performance change when applying our additional loss $\mathcal{L}_{QE}$ to the LED SOTA model on QAsper.} %
\label{tab:qasper_results}
\end{table*}

\begin{table}[t]
\centering
\footnotesize
\setlength{\tabcolsep}{7pt} 
        \begin{tabular}[b]{@{}clrrr@{}}
            \toprule
            &Model            & Ans & Sup & Joint \\ \midrule
       \multirow{4}{*}{\rotatebox[origin=c]{90}{base size}}
            &Longformer-base \citeyearpar{longformer}  & 74.5   & 83.9    & 64.5     \\ 
            &+$\mathcal{L}_{QE}$ ($\Delta$)  & +0.9  & +0.2    & +1.0    \\ 
            \cmidrule(lr){2-5}
            &\textsc{CDLM}-base \citeyearpar{caciularu-etal-2021-cdlm-cross} & 74.7 & 86.3 & 66.3\\
            &+$\mathcal{L}_{QE}$ ($\Delta$)  & +0.9 & +0.3 & +0.2\\
            \midrule
            \multirow{4}{*}{\rotatebox[origin=c]{90}{large size}}
            &Longformer-large \citeyearpar{longformer} & 81.3   & 88.3    & 73.2     \\ 
            &+$\mathcal{L}_{QE}$ ($\Delta$) & +0.3   & +0.1    & +0.9     \\ 
            \cmidrule(lr){2-5}
            &\textsc{CDLM}-large \citeyearpar{caciularu-etal-2021-cdlm-cross}& 81.3 & 89.1 & 73.8\\
            &+$\mathcal{L}_{QE}$ ($\Delta$)& +0.3 & +0.6 & +0.8\\
            \bottomrule
        \end{tabular}
\caption{HotpotQA-distractor results ($F_1$) for the dev set. We use the ``base'' and ``large'' model size results of CDLM and the Longformer for direct comparison. Ans: answer span, Sup: Supporting facts.} 
\label{tab:hotpotqa}
\end{table}
\begin{table}[!tb]
    \newcommand{\colindent}{\;}
    \centering
    \resizebox{0.49\textwidth}{!}{
    \begin{tabular}{@{}lrr@{}}
    \toprule
    \phantom{fwidsvhckzxjvchndfzxvvgdaczfc} & Joint & $\Delta$\\
    \midrule
    \textsc{CDLM}-large + $\mathcal{L}_{QE}$ and $sim(\cdot)$ is Eq. \ref{eq:cos_k} & 74.6 & \\
    \colindent $-$ using a single $\tau$ parameter for all the question types & 74.2 & -0.4 \\
    \midrule
    \colindent $-$ $sim(\cdot)$ is the dot product & 73.1 & -1.5\\
    \colindent $-$ $sim(\cdot)$ is the cosine similarity & 74.0 & -0.6\\
    \colindent $-$ $sim_k(\cdot)$ is the bilinear distance (Eq.~\ref{eq:first_s}) & 73.7 & -0.9 \\
    \colindent $-$ $sim(\cdot)$ is Eq.~\ref{eq:cos_k} with a single projection ($W_k=W$) & 74.0 & -0.6 \\
        \midrule
    \colindent $-$ w/o incorrect question type projected negatives & 74.2 & -0.4 \\
    \bottomrule
    \end{tabular}}
    \caption{Similarity function ablation results (Joint $F_1$) of CDLM-large and our loss term on the HotpotQA-distractor dev set.}
    \label{tab:ablations}
\end{table}
\section{Evaluation and Analysis}
In this section, we provide details about the experiments that we conducted and their outcomes. 
\subsection{Experimental Setup}

In order to demonstrate the contribution of our method, we evaluated it over the recent QAsper dataset~\cite{dasigi-etal-2021-dataset} and the well-known HotpoQA dataset~\cite{yang-etal-2018-hotpotqa},
which share the same setup (\sect \ref{sec:background}). 

\paragraph{QAsper}~\cite{dasigi-etal-2021-dataset} is a long-document QA dataset which was built over academic papers, where NLP practitioners were recruited to (abstractedly) generate questions following the title and the abstract of a particular paper, as well as creating the the correct evidence and answers to those questions based on the entire paper. More than half of the examples in QAsper require collecting evidences from multiple evidences in the given paper. For this benchmark, $\mathcal{L}_{QA}$ represents the sum of the teacher-forced text generation and evidence classification loss functions, in a multi-task training setup.

\paragraph{HotpotQA}~\cite{yang-etal-2018-hotpotqa} introduced the task of multihop extractive question answering, in the reading comprehension setting, where the inputs are a question and multiple paragraphs from various related and non-related documents. A model is queried to extract answer spans and evidence sentences, where it should handle challenging questions that require finding and reasoning over multiple supporting documents. For the models we applied to this benchmark, $\mathcal{L}_{QA}$ represents the standard cross-entropy answer extraction loss.

To test the contribution of our $\mathcal{L}_{QE}$ loss, We replicated the experiments described in~\citet{dasigi-etal-2021-dataset} and~\citet{caciularu-etal-2021-cdlm-cross} for QAsper and HotpotQA, respectively. For QAsper, we finetuned the LED-base model,\footnote{According to~\citet{dasigi-etal-2021-dataset}, LED-base outperforms LED-large over QAsper.} and evaluated it on the question answering and evidence selection tasks. For HotpotQA, we used the Longformer model and CDLM\footnote{CDLM was shown to be an effective long-range cross-encoder model for HotpotQA.} as the backbone long-sequence language models for this task. Since CDLM was provided only as a base-sized model, we pretrained a larger version of the CDLM model, and hence used both the base and large versions of both Longformer and CDLM. For further details see Appendix~\ref{subsec:qasper} and~\ref{subsec:hotpotqa}.

For both benchmarks, we performed a grid search for determining the hyper-parameters of the contrastive loss (more details in Appendix~\ref{sec:contrastive}).
\subsection{Results and Analysis}
\paragraph{Qustion-Evidence Similarity Analysis.} As a preliminary assessment, we first investigate the question and evidences representations of models trained on the HopotQA dataset. We motivate the use of our method by presenting the mean Average Precision (mAP) ranking results produced according to the question-sentence cosine similarities for the marker tokens trained representations of the CDLM-large model. \red{From Table~\ref{tab:map}, we observe that without applying our additional loss term, the question representations are distant from the evidence representations. Using a single learned projection increases this desirable similarity, and using a learned projection per question type yields the highest mAP scores. Hence, integrating question-type aware linear projections can be a beneficial part of our contrastive loss, and overall it further improves the QA results as we show next.} An additional illustration of this effect, where we visualize the marker representations, appears in Appendix~\ref{sec:appndx1}.

\paragraph{Main results.} We adopted the F1 evaluation metrics corresponding to the original works~\cite{dasigi-etal-2021-dataset,longformer}. Tables \ref{tab:qasper_results} and \ref{tab:hotpotqa} present the evaluation results over the QAsper and HotpoQA datasets, respectively. We show the performance difference when adding our additional loss term with ``+$\mathcal{L}_{QE}$'' (and question-type similarity function from Eq.~\ref{eq:cos_k}). 

As shown in the table, the addition of $\mathcal{L}_{QE}$ exhibits the best performance across all examined models and benchmarks, clearly demonstrating its consistent advantage. Note that maximizing the question-evidence similarity resulted also in evidence detection improvement -- see the ``Evidence'' and the ``Sup'' metrics in Tables~\ref{tab:qasper_results} and ~\ref{tab:hotpotqa}, respectively. All the results are statistically significant using the bootstrap test with ${p<0.01}$~\cite{dror-etal-2018-hitchhikers}.

\paragraph{Ablations.} Table~\ref{tab:ablations} demonstrates ablation study results for evaluating the effectiveness of our design decisions. Using a constant temperature parameter for all question types, as well as using different degenerated similarity functions, exhibits lower performance. Further, the last row in Table~\ref{tab:ablations} shows that treating correct answers that are projected to the wrong question type as negatives also improves the results.
Overall, the ablation study shows the advantage of using Eq.~\ref{eq:cos_k} as a similarity function that provides fine-grained expressive modeling for each question type, in its own sub-space. 

\paragraph{Discussion.} An additional theoretical justification to our contrastive learning is provided in~\cite{gao-etal-2021-simcse}, where we can imply that our loss term improves the uniformity and therefore the expressiveness of the question and evidence representations. Moreover, one can attribute the success of our contrastive loss to the fact that long-range transformer models lack long-range signals during pretraining, and hence such explicit modeling as we suggest is necessary. In fact, comparing CDLM's results to the Longformer illustrates that our cotrastive term has higher impact on models without global attention-based pretraining.

\section{Conclusion}
In this work, we proposed an effective sequence-level contrastive loss for improving the performance of long-range transformers in solving QA tasks that require reasoning over long contexts. We demonstrate consistent improvement when using our approach on three different models over two different benchmarks.
For future work, we propose exploring variations of our proposed supervised loss on other long-context tasks, such as long-document and multi-document summarization, and integrating our method into information retrieval re-ranker models.

\section*{Acknowledgments}
We thank the BIU-NLP lab and the Semantic Scholar research team at AI2 for fruitful discussions and helpful feedback.
The work described herein was supported by the PBC fellowship for outstanding PhD candidates in data science, in part by by the Israel Science Foundation grant 2827/21, and by a grant from the Israel Ministry of Science and Technology.
\section*{Ethical Considerations}
Our work in understanding the role of maximizing the similarity between question and evidence pairs. Therefore, there is a limited risk associated with the quality of annotated evidence sentences in the dataset, as there is no guarantee that our model will always
generate non-biased and factual content. Therefore, caution must
be exercised when the model is deployed in practical settings, where the evidence quality is vague and cannot be verified.

\bibliography{acl2021,anthology}
\bibliographystyle{acl_natbib}
\clearpage
\appendix
\section*{Appendix}

\section{Question-Evidence Similarity Demonstration}
\label{sec:appndx1}

In this section, we further interpret the outcome of our contrastive loss.

\begin{figure*}
\centering
\includegraphics[width=.98\linewidth]{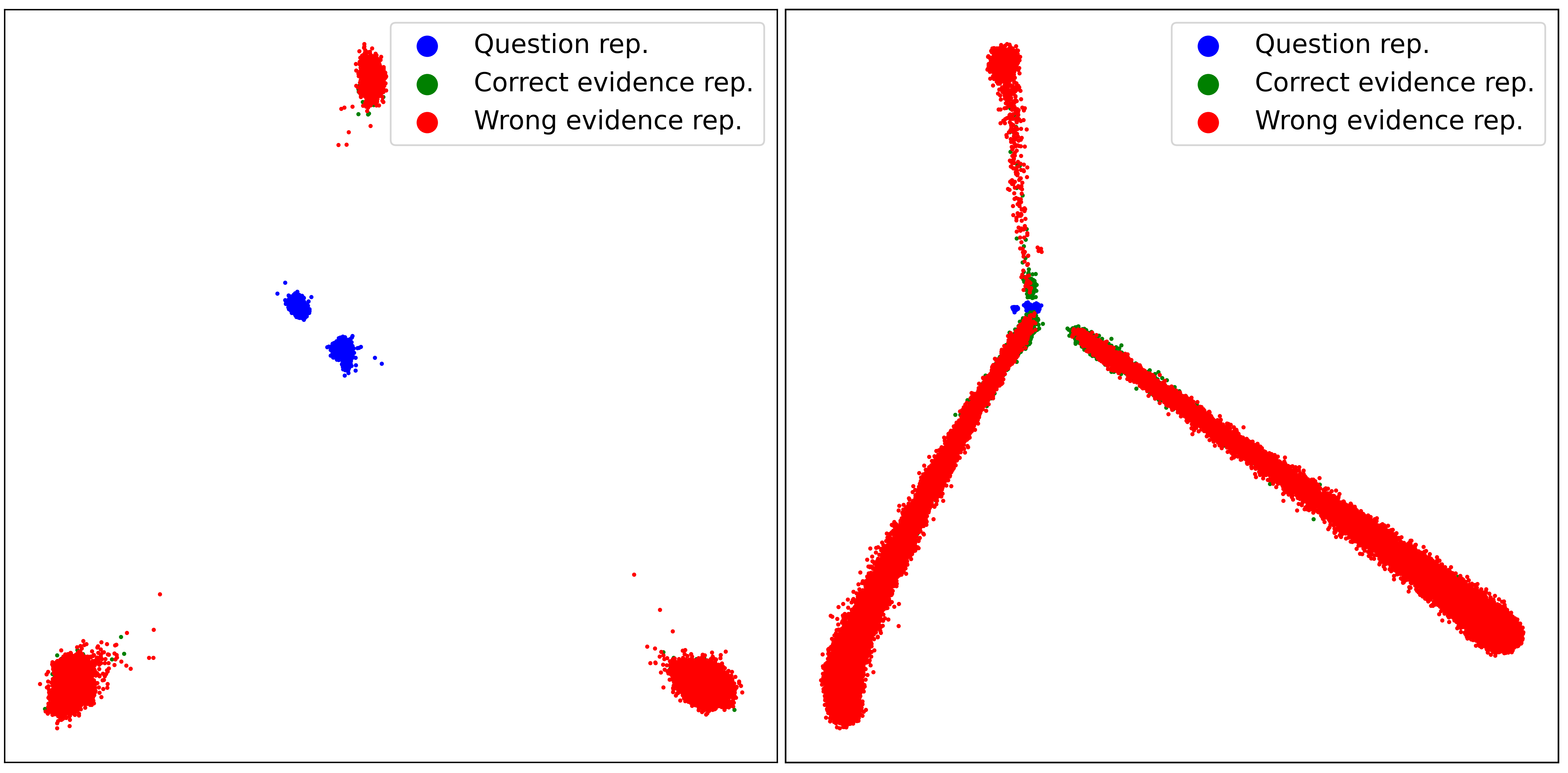}
  \caption{PCA plots of the learned question and answer token embeddings on the HotpotQA validation set, comparing early training epochs results (left) and results after model convergence (right). The wrong evidence representations correspond to both wrong evidences, or correct evidences using the wrong question type projections (our soft negatives).}
  \label{fig:pca}
\end{figure*}

We apply PCA over the relevant normalized token representations of the validation data of HotpotQA (i.e., the question and answers representations in Fig.~\ref{fig:model}), and depicted them in~\ref{fig:pca}. The projected representations of the correct and wrong answers are equally distributed at the beginning of the training (left figure). After several epochs when the model converged (right figure), the answer representations' manifold got closer to the questions' representations (in terms of radial distance). Each beam in the figure corresponds to a different question type (there are 3 in HotpotQA). The correct evidence representations (green dots) are the closest among the whole answer representations, confirming that our additional contrastive loss term generalizes and maximizes the question-evidence similarity.

\section{Datasets and Finetuning Details}
\label{sec:appndx2}
In this section, we provide details, regrading finetuning and hyper-parameter configuration, over the benchmarks we used during our experiments.

\subsection{QAsper}
\label{subsec:qasper}
Since some of the questions included in QAsper are not answerable, we apply our contrastive loss only over examples that are answerable and contain at least one evidence sentence. 

We train all models using the Adam optimizer~\citep{kingma2014adam} and a triangular learning rate scheduler \cite{howard2018universal} with 10\% warmup. To determine number of epochs, peak learning rate, and batch size, we performed manual hyperparameter search on a subset of the training data. We searched over \{1, 3, 5\} epochs with learning rates \{$1e^{-5}$, $3e^{-5}$, $5e^{-5}$, $9e^{-5}$\}, and found that smaller batch sizes generally work better than larger ones. Our final configuration was 10 epochs, peak learning rate of $5e^{-5}$, and batch size of 2, which we used for all reported experimental settings.  When handling full text, we use gradient checkpointing~\cite{gradckpt} to reduce memory consumption. We run our experiments on a single RTX 8000 GPU, and each experiment takes 30--60 minutes per epoch.

\subsection{HotpotQA}
\label{subsec:hotpotqa}
We used the HotpotQA-distractor dataset~\cite{yang-etal-2018-hotpotqa}. Each example in the dataset is includes a question and 10 paragraphs from different documents, extracted from Wikipedia. Two gold paragraphs include the relevant information for properly answering the question, mixed and shuffled with eight distractor paragraphs (for the full dataset statistics, see~\citet{yang-etal-2018-hotpotqa}). There are two goals for this task: detecting the supporting facts, i.e., evidence sentences, as well as extraction of the correct answer span. 

For preparing the data for training and evaluation, we follow the same finetuning scheme of the CDLM~\cite{caciularu-etal-2021-cdlm-cross} and the Longformer~\cite{longformer}; For each example, we concatenate the question and all the 10 paragraphs in one long context. We particularly use the following input format with special tokens and our document separators: ``\texttt{[CLS] [q] question [/q] $\langle$\texttt{doc-s}$\rangle$$\langle$t$\rangle$ $\texttt{title}_{\texttt{1}}$ $\langle$/t$\rangle$} \texttt{$\langle$s$\rangle$} $\texttt{sent}_{\texttt{1,1}}$ \texttt{$\langle$/s$\rangle$} \texttt{$\langle$s$\rangle$} $\texttt{sent}_{\texttt{1,2}}$ \texttt{$\langle$/s$\rangle$} $\langle$\texttt{/doc-s}$\rangle$ \texttt{...} \texttt{$\langle$t$\rangle$ $\langle$\texttt{doc-s}$\rangle$ $\texttt{title}_{\texttt{2}}$ $\langle$/t$\rangle$ }  $\texttt{sent}_{\texttt{2,1}}$ \texttt{$\langle$/s$\rangle$} \texttt{$\langle$s$\rangle$} $\texttt{sent}_{\texttt{2,2}}$ \texttt{$\langle$/s$\rangle$} \texttt{$\langle$s$\rangle$} \texttt{...}'' where \texttt{[q]}, \texttt{[/q]}, $\langle$\texttt{t}$\rangle$, $\langle$\texttt{/t}$\rangle$, \texttt{$\langle$s$\rangle$}, \texttt{$\langle$/s$\rangle$}, \texttt{[p]} are special tokens representing, question start and end, paragraph title start and end, and sentence start and end, respectively. The new special tokens were added to the models vocabulary and randomly initialized before task finetuning. We use global attention to question tokens, paragraph title start tokens as well as sentence tokens. 
The model's structure is taken from ~\citet{longformer}.

As in~\citet{longformer,caciularu-etal-2021-cdlm-cross}, we finetune our models for 5 epochs, using a batch size of 32, learning rate of $1e^{-4}$, 100 warmup steps. Finetuning on our models took $\sim$6 hours per epoch, using four 48GB RTX8000 GPUs for finetuning our models. For generating the CDLM-large results, we pretrined our version using the code from \url{https://github.com/aviclu/CDLM/tree/main/pretraining}.

\section{Contrastive Loss Details}
\label{sec:contrastive}
In this section, we provide the details for reproducing our contrastive term, which is relevant for both QAsper and HotpotQA.

We searched over $d\times\{d,\frac{d}{2},\frac{d}{4},\frac{d}{8}\}$ to determine the linear projections' dimensions, where $d$ is the model's hidden layer representation dimension (it depends on the size of the model). In order to determine the temperature hyperparameter $\tau$, we searched over $\{0.2,0.4,0.6,0.8,1.0\}$ per question type (if applicable). We also applied dropout with a rate of ${p=0.1}$ over the linear projections, which consistently improved the results over all the benchmarks. Finally, we searched for the best performing $\lambda$ hyperparameter over the values of $\{0.2,0.4,0.6,0.8,1.0\}$.

\end{document}